\documentclass[runningheads]{llncs}
\UseRawInputEncoding
\usepackage{graphicx}
\usepackage{cite}
\usepackage{booktabs}
\usepackage{algorithm,algorithmic}
\usepackage{hyperref}
\usepackage{subcaption}
\usepackage{amssymb}
\usepackage{pifont}
\usepackage{xcolor}

\definecolor{pink}{RGB}{219, 48, 122}

\begin{document}

\title{ViPTT-Net: Video pretraining of spatio-temporal model for tuberculosis type classification from chest CT scans}
\titlerunning{ViPTT-Net: Video pretraining for tuberculosis type classification}

\author{Hasib Zunair\inst{1} \and
Aimon Rahman\inst{2} \and
Nabeel Mohammed\inst{3}}
%
%
\institute{Concordia University\inst{1} and North South University\inst{2,3}\\
\email{hasibzunair@gmail.com}\inst{1},
\email{aimon.rahman@northsouth.edu}\inst{2}, \email{nabeel.mohammed@northsouth.edu}\inst{3}
}
\maketitle              
\begin{abstract}
Pretraining has sparked groundswell of interest in deep learning workflows to learn from limited data and improve generalization. While this is common for 2D image classification tasks, its application to 3D medical imaging tasks like chest CT interpretation is limited. We explore the idea of whether pretraining a model on realistic videos could improve performance rather than training the model from scratch, intended for tuberculosis type classification from chest CT scans. To incorporate both spatial and temporal features, we develop a hybrid convolutional neural network (CNN) and recurrent neural network (RNN) model, where the features are extracted from each axial slice of the CT scan by a CNN, these \textit{sequence of image features} are input to a RNN for classification of the CT scan. Our model termed as ViPTT-Net, was trained on over 1300 video clips with labels of human activities, and then fine-tuned on chest CT scans with labels of tuberculosis type. We find that pretraining the model on videos lead to better representations and significantly improved model validation performance from a kappa score of 0.17 to 0.35, especially for under-represented class samples. Our best method achieved 2nd place in the ImageCLEF 2021 Tuberculosis - TBT classification task with a kappa score of 0.20 on the final test set with only image information (without using clinical meta-data). All codes and models are made available.~\footnote{\href{https://github.com/hasibzunair/viptt-net}{\texttt{\textcolor{pink}{https://github.com/hasibzunair/viptt-net}}}}

\end{abstract}

{\let\thefootnote\relax\footnotetext{Copyright \textcopyright\ 2021 for this paper by its authors. Use permitted under Creative Commons License Attribution 4.0 International (CC BY 4.0). CLEF 2021 -- Conference and Labs of the Evaluation Forum, September 21--24, 2021, Bucharest, Romania}}

\section{Introduction}

Tuberculosis (TB) is a potentially fatal disease that generally affects the lungs. The disease spreads through cough, spit or sneeze and can remain latent within the human body. Although X-rays and microscopic analysis of bodily fluid are generally used to diagnose the disease, Computed Tomography (CT) provides detailed information about the infection. Deep learning models demonstrate promising results in diagnosing TB from both X-rays and CT scans \cite{lakhani2017deep, rajpurkar2020chexaid, nash2020deep, li2020deep}. These methods are also proven to be effective for severity scoring of the infection as well \cite{zunair2020uniformizing, zunair2019estimating, biomed}. However, the types of TB can vary and that may require a different course of treatment, making the identification of TB type an important real-life problem. The deep learning models are yet to provide high accuracy in this particular task. 

The most challenging part of working with CT scan data is, it is three-dimensional (3D), meaning along with height and width, each data point contains depth information. Processing 3D data can be very computationally expensive and may require pre-processing before training a model. These pre-processing steps may include choosing selective slices or transforming slices into uniform sizes. Both of these techniques may contribute to losing some depth information. One of the most effective method of processing 3D data is to resize it into fixed dimension using spline interpolation along x, y and z- axis which was demonstrated in \cite{zunair2020uniformizing}. Although the method showed promising result in TB severity scoring, might not perform well on other more sophisticated tasks such as classifying TB types.

An alternative to directly processing 3D data is to decompose each scan into individual slices and afterward feeding them 2D CNN \cite{ref_7,ref_8,UCSD,21-ref}. The problem with this method is, as the slices are considered independent to each other, the spatial information along the z-axis is missing during training. Moreover, in this method, the label for the whole volume is assigned to each individual slice, which might not be the case when it comes to CT scans. More specifically, a CT scan of infected lungs may contain uninfected 2D slices.

In this work, we build a model to predict multiple tuberculosis types from chest CT scans. We pretrain a hybrid convolutional neural network (CNN) and recurrent neural network (RNN) model on human action recognition task, and the fine-tune the model for the tuberculosis type classification task. Pretraining significantly improves performance, especially for minority classes. The method is evaluated on the Image-CLEF 2021 Tuberculosis - TBT classification task which achieves 2nd place~\footnote{\href{https://www.aicrowd.com/challenges/imageclef-2021-tuberculosis-tbt-classification/leaderboards}{\texttt{\textcolor{black}{imageclef-2021-tuberculosis-tbt-classification/leaderboards}}}} with a kappa score of 0.20 and accuracy of 0.42.

We summarize our contributions as follows:
\begin{enumerate}
  \item We pretrain a hybrid CNN-RNN model, termed ViPTT-Net, on human action recognition task, and fine-tune the model on a small dataset of CT scans with labels indicating tuberculosis types.
  \item We show pretraining ViPTT-Net on realistic videos improve performance for tuberculosis type classification, especially for under-represented class samples.
  \item We evaluate our best method on the Image-CLEF 2021 Tuberculosis - TBT classification task which achieves 2nd place overall.
\end{enumerate}

\section{Methodology}

We start by problem formulation followed by presenting the main building blocks of our proposed method for multi-class CT image classification. 

\subsection{Problem Formulation}
Given the labels of a set of CT scans, the objective is to predict the unknown labels of the new CT scans~\footnote{\href{https://www.imageclef.org/2021/medical/tuberculosis}{\texttt{\textcolor{black}{https://www.imageclef.org/2021/medical/tuberculosis}}}}. More specifically, our goal is to learn a discriminative function $f(\textbf{X}) \in  \{\textbf{1},\textbf{2}, \textbf{3},\textbf{4},\textbf{5}\}$, where the numbers represent the tuberculosis type: \textit{Infiltrative}, \textit{Focal}, \textit{Tuberculoma}, \textit{Miliary} and \textit{Fibro-cavernous} repsectively. $\textbf{X}$ represents a CT scan volume of size $D\times W\times H$, where D, W, and H represent the depth, width, and height of the volume respectively. The task is considered as a multi-class volumetric image classification problem. A 3D volumetric scan $\textbf{X}$ can also be viewed as a time-series of 2D slices $\{{X_1,\dots, X_D\}}$. Therefore, we can also frame the task as a time-series sequence classification problem. 

\subsection{ViPTT-Net Model}
While convolutional neural networks (CNNs) have shown promising results at processing image data~\cite{lecun1998gradient,krizhevsky2012imagenet,he2016deep}, the same can be said for recurrent neural networks (RNN) for sequential data~\cite{ebrahimi2015recurrent,salehinejad2017recent,salehinejad2020deep}. We developed a hybrid convolutional neural network (CNN) and recurrent neural network (RNN) model termed ViPTT-Net which is capable of incorporating both spatial and temporal features in the learning process. ViPTT-Net takes as input a CT scan and outputs probability predictions which indicates the type of tuberculosis in that CT scan. To deal with arbitrary volumes, we first resize the CT scan to a fixed size $70\times 224\times 224$ using spline interpolated zoom (SIZ)~\cite{zunair2020uniformizing} which exploits the full geometry of the 3D volume by interpolating over the z-axis.  

\medskip\noindent\textbf{Learning spatial features using CNN.}\quad To learn spatial features, ViPTT-Net consists of a VGG-16 model as the feature extractor which is pretrained on ImageNet~\cite{deng2009imagenet}. The VGG-16 model has 16 layers with learnable weights: 13 convolutional layers, and 3 fully connected layers~\cite{simonyan2014very}. We extract features from the last convolutional layer which results in a 512 dimensional feature vector for a 2D input image. Since a CT scan consists of multiple 2D axial slices (70 in our case), we wrap the VGG-16 model in a time-distributed layer which is then applied to every axial slice (temporal frame) of the CT scan independently. The final output is a \textit{sequence of image features} where the sequence length is 70 and each of the sequences are a 512 dimensional feature vector. 

As the VGG-16 feature extractor accepts inputs of 3-channels, we map the 1-channel axial slices of the CT scan slices to 3-channel using a convolutional layer with 3 filters and kernel size of $1\times 1\times 1$ before input to the feature extractor.

\medskip\noindent\textbf{Learning temporal features using RNN.}\quad Following the above, the sequence of image features are aggregated using an RNN. RNNs are a type of neural network which transforms a sequence of inputs into a sequence of outputs. Using an RNN, the temporal order of the axial slices is preserved. We use the long short-term memory (LSTM)~\cite{greff2016lstm} with 256 units and tanh activation. It is important to mention that since we are interested in classifying the full sequence, the output from the LSTM results in a 256 dimensional aggregated feature vector from the last time step coming from the full sequence of image features, rather than producing a sequence of outputs. We also experimented with gated recurrent unit (GRU) but did not get good results on the validation set. 

\begin{figure}
\centering
\includegraphics[width=0.95\textwidth]{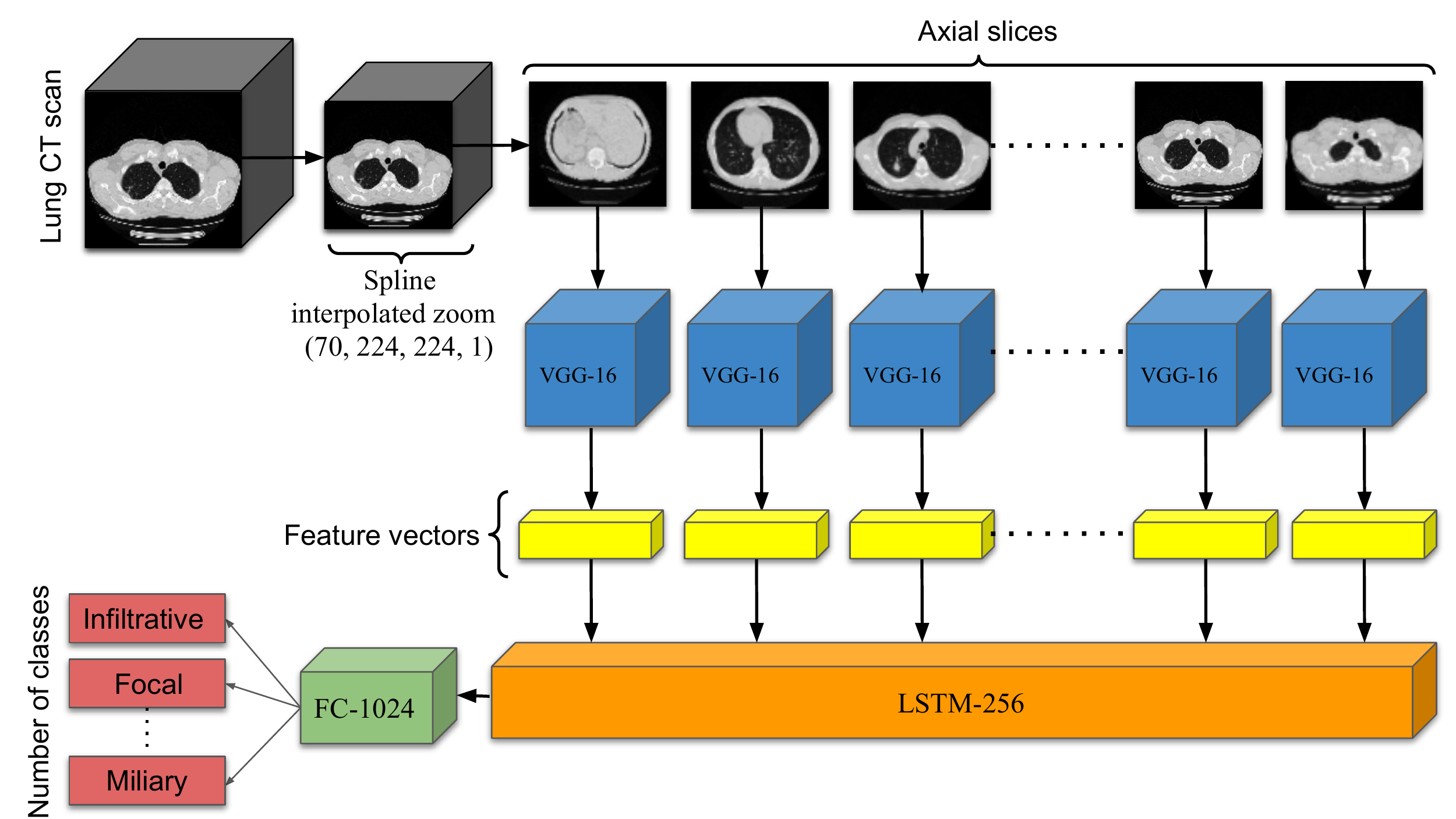}
\vspace{1pt}
\caption{\small Schematic layout of the hybrid CNN-RNN model ViPTT-Net. Given a 3D CT scan of arbitrary size, uniform resizing is performed across all dimensions using SIZ~\cite{zunair2020uniformizing}. Features are extracted from all the axial slices of the processed CT scan to output a sequence of image features using a VGG-16 model. These sequence of image features are input to an LSTM layer followed by dense layers of 1024 neurons and finally 5 with softmax activation for the multi-class classification problem.}
\label{model}
\end{figure}

Finally, the aggregated 256 dimensional feature vector is passed to a dense layer with 1024 units with rectified  linear  hidden  units  (ReLUs) activation and a dense layer with a softmax function (i.e. a dense softmax layer of 5 units for the multi-class classification case) that yields the probabilities of predicted classes. While training on UCF50, instead of a softmax with 5 units, we use a softmax with 10 units for the 10 classes. An illustration of ViPTT-Net is shown in Fig~\ref{model}.

\subsection{Pretraining on Human Action Recognition Task}
\begin{figure}
\begin{subfigure}{.45\textwidth}
  \centering
  \includegraphics[width=1.0\linewidth]{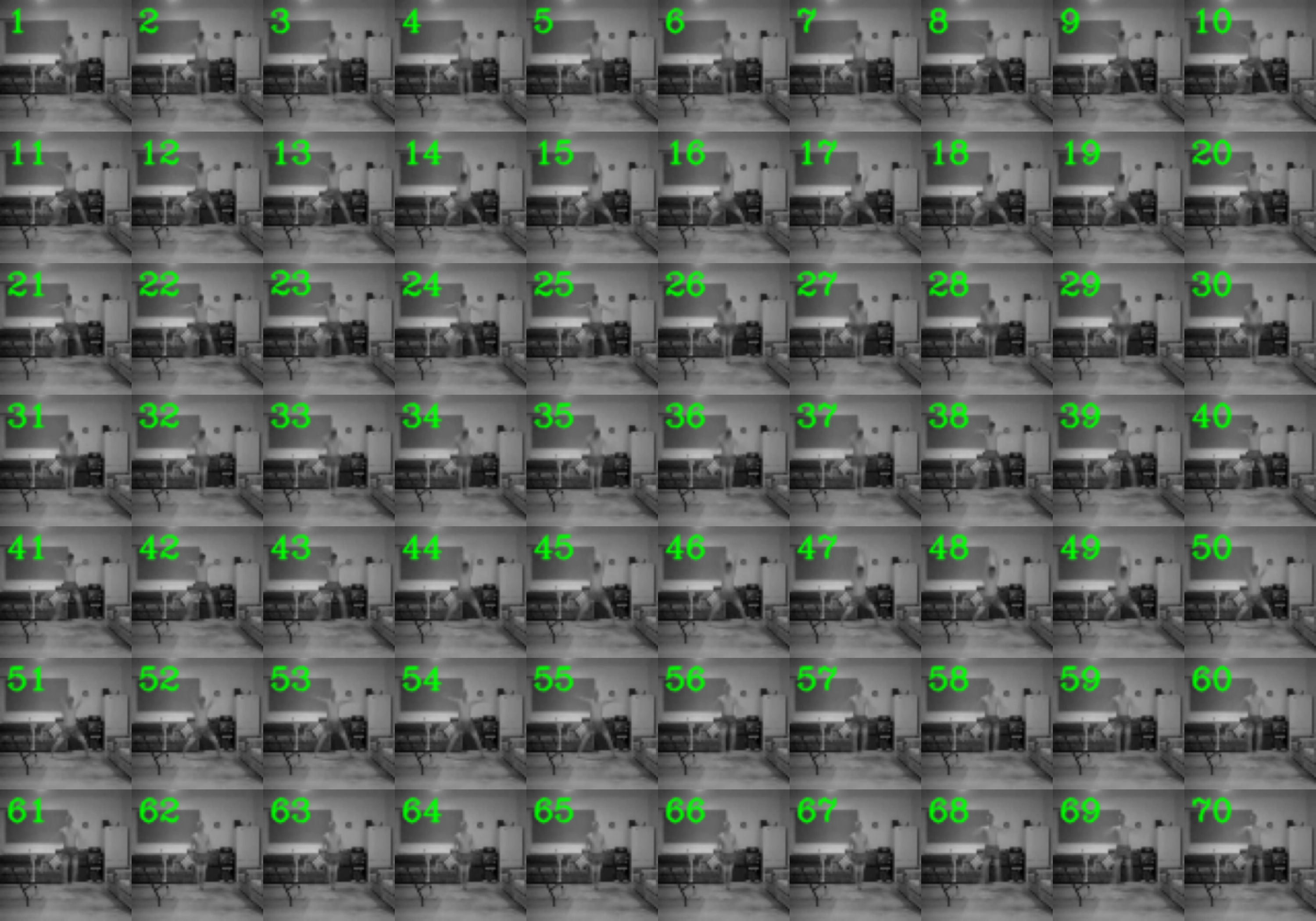}
  \caption{Sequence of images from a video of a man playing~\textit{Jumping Jack}.}
  \label{fig:1a}
\end{subfigure}%
\hspace{0.2cm}
\begin{subfigure}{.45\textwidth}
  \centering
  \includegraphics[width=1.0\linewidth]{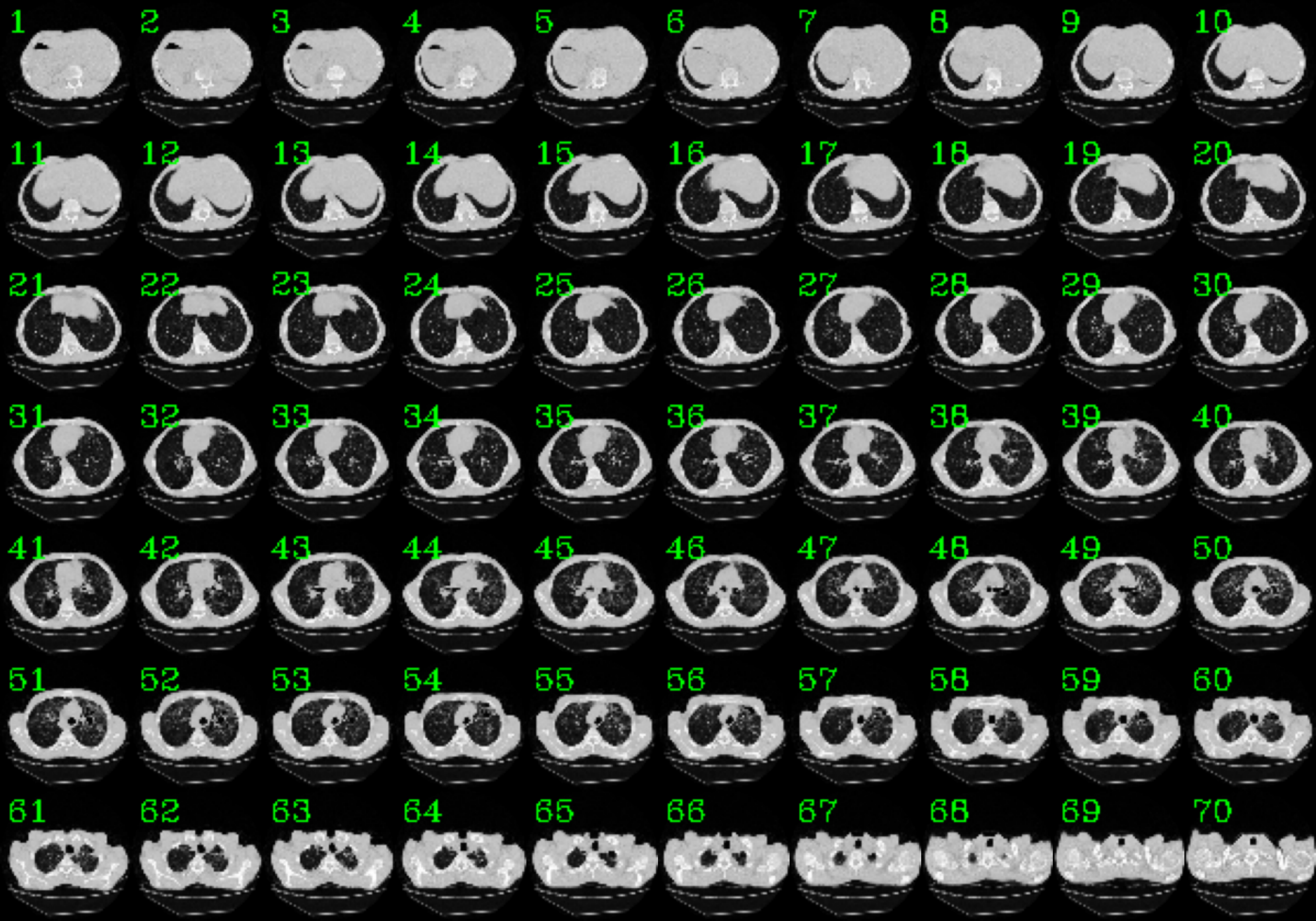}
  \caption{Sequence of axial images from a CT scan with the label~\textit{Infiltrative}.}
  \label{fig:1b}
\end{subfigure}
\caption{Illustration of image sequences of a video and a CT scan from the UCF50 and ImageCLEF 2021 - TBT datasets. Both samples are resized to depth of 70.}
\label{fig:images}
\end{figure}

ViPTT-Net was first pretrained on a subset of the UCF50 dataset~\cite{reddy2013recognizing}. UCF50 is human action recognition dataset with 50 action categories, consisting of realistic videos taken from youtube. The classes of different action are Baseball Pitch, Jumping Jack, Kayaking \textit{etc}. Due to compute constraints, we use 1366 videos from 10 classes selected randomly: Mixing, Tennis Swing, Horse Riding, Jump Rope, Jumping Jack, Baseball Pitch, Rowing, SkateBoarding, Walking With Dog, Skijet. Video clips are resized to $70\times 224\times 224$, where 70 is the number of temporal frames (image sequences), and 224 is width and height of the temporal frame. It is also important to mention that each RGB frame of the video is converted to grayscale (single channel) since the CT scans also have single channel values. An instance of an action video is shown in Fig~\ref{fig:1a}.

Similar to pretraining for 2D image problems~\cite{zhuang2020comprehensive,zunair2020melanoma}, after training ViPTT-Net on a subset of UCF50 dataset, the network was fine-tuned on the tuberculosis type classification task by replacing the final fully connected 10-way softmax layer with a 5-way softmax. During training ViPTT-Net on UCF50, the weights of the VGG-16 feature extractor was frozen. And while fine-tuning on the tuberculosis type classification task, all the network layers were trained.

\subsection{Weighted Loss and Data Augmentation}

Due to heavy class imbalance, we assign weights in the loss function for each tuberculosis type with the goal to reduce biasness towards the over-represented class samples. Prior to training using weighted loss, the weights are computed over the training set. 

It is a standard practice to perform data augmentation to improve generalization, especially when there are limited number of training data. Data augmentation basically creates modified versions of the input data in a dataset through random transformations such as horizontal and vertical flip, zoom augmentation, horizontal and vertical shift, \textit{etc}. While training, the 3D CT scans are rotated with degree of rotations picked randomly from $[-20, -10, -5, 0, 5, 10, 20]$ as a form of data augmentation. Notice that we added 0 in the range which means that the model looks at both augmented and non-augmented data. We experimented without adding 0 in the range and led to poor results on the validation set. We also tried blur and random shifts but did not get good results.

\section{Experiment Setup}
We describe experimental details, \textit{i.e.}, dataset,  implementation details, evaluaton metrics, \textit{etc.},  and present quantitative results, comparing different training strategies using ViPTT-Net.

\subsection{Datasets and Preprocessing}
The dataset is provided by ImageCLEF Tuberculosis Type 2021 Challenge~\cite{ImageCLEF2021,tub21}. It contains CT scans of a total of 1338 TB patients, 917 of them have been used for training and 421 for the test set. Each scan has a label that indicates one of the TB types- Infiltrative, Focal, Tuberculoma, Miliary and Fibro-cavernous. Each CT scan belongs to only one patient. The images have dimensions of 512 x 512 pixels with varying depth sizes. In addition to labels, some scans have additional meta-data. All the scans have auto-generated lungs mask, although these masks are generally missing in largely affected areas or have rough bounds. The original data is in NIFTI format, storing the raw
voxel intensity in Hounsfield units (HU). An instance of a CT scan with Infiltrative tuberculosis type is shown in Fig~\ref{fig:1b}.

\subsection{Implementation Details}
All experiments are performed on a Linux workstation running 4.8Hz and 64GB RAM with and RTX 3080 GPU. Experiments are conducted using Python programming language~\cite{van2007python}. ViPTT-Net is implemented in Keras~\cite{chollet2015keras}, with TensorFlow backend~\cite{abadi2016tensorflow}.

ViPTT-Net is trained end-to-end using stochastic gradient descent (SGD) optimization algorithm to minimize the categorical cross-entropy loss function with an initial learning rate of 0.001 and batch size of 2. A factor of 0.1 is used to reduce the learning rate once the loss stagnates. Training is continued until the validation loss stagnates using an early stopping mechanism, and then the best weights are retained. To keep consistent data proportions same and ensure reproducibility, we perform a stratified train and validation split with a ratio of 80/20 (732/184 CT scans) on the training data provided by ImageCLEF 2021 - TBT. 

Similar steps are followed while training on a subset of the UCF50 dataset, except we do not use any data augmentation in this case.

\subsection{Evaluation Metrics} \label{D}
According to challenge rules, the task is evaluated as a multi-class classification problem. The main evaluation metric is the kappa score. Kappa measures the inter-rater reliability for categorical items. 

\begin{equation}
    \kappa  \equiv \frac{p_0 - p_e}{1 - p_e}
\end{equation}
Here, $p_0$ indicates the relative observed agreement among raters and $p_e$ is the probability of agreement by chance. If raters are in complete agreement then $\kappa = 1$ and no agreement other than by chance will result in $\kappa = 0$. The value would be negative if there is no effective agreement among the raters or the agreement is worse than random. Additionally we also show accuracy (ACC) and per-class F1 scores. For all metrics, the higher the better.

\subsection{Results}

We study the effect of training ViPTT-Net for the task of tuberculosis type classification using different training strategies:

\medskip\noindent\textbf{No PT.}\quad ViPTT-Net is trained from scratch on the 732 CT scans annotated for tuberculosis types

\medskip\noindent\textbf{PT.}\quad ViPTT-Net is first pretrained on a subset of the UCF50 dataset which around 1300 video clips annotated for human activities. Then the last layer is replaced with a five unit softmax and fine-tuned on 732 CT scans annotated for tuberculosis types.

\medskip\noindent\textbf{PT+CW.}\quad Same as PT, but additionally weighted loss is used with weights computed over the training set of 732 CT scans annotated for tuberculosis types

\medskip\noindent\textbf{PT+CW+AUG.}\quad Same as PT+CW, but additionally the 3D volumes are randomly rotated during training.

Table~\ref{table:val} summarizes the results which show that pretraining ViPTT-Net on a subset of the UCF50 dataset followed by fine-tuning for tuberculosis type classification (PT) improves kappa score by 0.18 on the validation set compared to training ViPTT-Net from scratch (No PT). In both configurations (No PT and PT), F1 score is the lowest for Tuberculoma class with scores of 0 and 0.09 for No PT and PT respectively. This is improved by using weighted loss (PT+CW), where the model achieves Tuberculoma F1 score 0.37. The F1 score of PT+CW also improves for FibC class by a large margin compared to No PT, although there is a slight drop in overall kappa score. PT+CW with data augmentation (PT+CW+AUG) further improves performance, more specifically for Miliary class by a large margin.

\begin{table}[t]
\setlength{\tabcolsep}{4pt}
\centering
\caption{Overall Kappa score and per-class F1 score achieved by different methods on the validation dataset of tuberculosis type classification. FibC denotes \textit{Fibro-cavernous}. For all methods, the ViPTT-Net model is used.}
\begin{tabular}{l c c c c c c}
\toprule
\bfseries Method & \bfseries Kappa & \bfseries Infiltrative & \bfseries Focal & \bfseries Tuberculoma & \bfseries Miliary & \bfseries FibC\\
\midrule
No PT & 0.17 & 0.61 & 0.46 & 0.0 & 0.2 & 0.14\\
PT   & \textbf{0.35} & \textbf{0.68} &   \textbf{0.56} & 0.09 & 0.4 & 0.48\\
PT+CW   &   0.30 & 0.59   &  0.41 & \textbf{0.37} & 0.33  & \textbf{0.65}\\
PT+CW+AUG & 0.33 & 0.59 &  0.47 & 0.27 & \textbf{0.54} & 0.61\\
\bottomrule
\end{tabular}
\label{table:val}
\end{table}

We also report the final submission results of our best model on the test set by ImageCLEF 2021 - TBT classification task in Table~\ref{table:test}. We observe a similar trend in performance improvements on the test set when using weighted loss. Our best method ViPTT-Net which is trained with weighted cross entropy loss and data augmentation performs the best on the test set.

\begin{table}[t]
\setlength{\tabcolsep}{4pt}
\centering
\caption{Overall Kappa and Accuracy score achieved by different methods on the test set by ImageCLEF  2021  Tuberculosis - TBT classification.}
\begin{tabular}{l c c}
\toprule
\bfseries Method & \bfseries Kappa & \bfseries Accuracy \\
\midrule
PT & 0.13 & 42.30\\
PT+CW  & 0.14 & 38.50\\
PT+CW+AUG   & \textbf{0.20} & \textbf{42.30}\\
\bottomrule
\end{tabular}
\label{table:test}
\end{table}

\section{Discussion and Conclusion}
We address the problem of predicting tuberculosis types from 3D chest CT scans. We develop a hybrid CNN-RNN model, termed ViPTT-Net, which is capable of learning both spatial and temporal features of the CT scan. Our experiments demonstrate that video pretraining on human action recognition task significantly improves performance of tuberculosis type classification rather than training the model from scratch. This is most significant for Miliary and Fibro-cavernous types without specifically aiming to improve performance for those classes. Interestingly, these classes along with Tuberculoma are the tuberculosis types which have the least amount of samples in the dataset. To further deal with class imbalance, we use a weighted loss function with weights computed over the training set that improves performance of under-represented classes significantly. Data augmentation also improved performance for few tuberculosis types on the validation set. On the test set, the highest kappa score was observed when pretraining ViPTT-Net on videos, using weighted loss function and data augmentation. This method achieved 2nd place in the ImageCLEF 2021  Tuberculosis - TBT classification task which operates on the CT image alone without using the additional patient meta-data and the lung segmentation masks.

%
%
%
\bibliographystyle{splncs}
\bibliography{paper}

\end{document}